\newif\ifcustomclass
\customclasstrue
\ifcustomclass
    \documentclass[10pt, a4paper, logo, twocolumn]{rai}
    \usepackage[toc,page]{appendix}
\else
    \documentclass[letterpaper, 10 pt, conference]{ieeeconf}  

    \IEEEoverridecommandlockouts                              
    
\fi

\usepackage{graphics} 
\usepackage{epsfig} 
\usepackage{times} 
\usepackage{amsmath} 
\usepackage{amssymb}  

\usepackage{cite}
\usepackage{caption}
\usepackage{subcaption}
\usepackage[ruled, linesnumbered]{algorithm2e}
\usepackage{pdfpages}
\usepackage[font=small,labelfont=bf]{caption}
\usepackage{array}
\usepackage{booktabs}
\usepackage{diagbox}
\usepackage{stfloats}
\usepackage{todonotes}
\usepackage{graphicx}
\usepackage{bm}
\usepackage[dvipsnames]{xcolor}

\SetAlFnt{\small}

\newcounter{methodstep}

\SetCommentSty{mycommfont}
\SetKwInput{KwInput}{Input}                
\SetKwInput{KwOutput}{Output}              
\makeatletter
\let\NAT@parse\undefined
\makeatother
\hypersetup{
     colorlinks = true,
     linkcolor = green,
     anchorcolor = green,
     citecolor = green,
     filecolor = green,
     urlcolor = cyan
 }
\usepackage{multirow}
\usepackage{caption}

\newcommand{\rv}[1]{\textcolor{black}{\textbf{} #1}}

\newcommand{\PaperName}{EMPM}

\makeatletter
\providecommand{\runningtitlefont}{}
\providecommand{\@runningtitle}{}
\makeatother

\begin{document}

\title{\LARGE \bf EMPM: Embodied MPM for Modeling and Simulation of Deformable Objects}

\author{
Yunuo Chen$^{1,2*}$,
Yafei Hu$^{1*}$,
Lingfeng Sun$^{1}$,
Tushar Kusnur$^{1}$,
Laura Herlant$^{1}$,
Chenfanfu Jiang$^{2}$
\\  $^{*}$equal contribution
\\
$^{1}$Robotics and AI Institute $^{2}$UCLA
}


\begin{abstract}
Modeling deformable objects -- especially continuum materials -- in a way that is physically plausible, generalizable, and data-efficient remains challenging across 3D vision, graphics, and robotic manipulation. Many existing methods oversimplify the rich dynamics of deformable objects or require large training sets, which often limits generalization. 
We introduce embodied MPM (EMPM), a deformable object modeling and simulation framework built on a differentiable Material Point Method (MPM) simulator that captures the dynamics of challenging materials. From multi-view RGB-D videos, our approach reconstructs geometry and appearance, then uses an MPM physics engine to simulate object behavior by minimizing the mismatch between predicted and observed visual data. 
We further optimize MPM parameters online using sensory feedback, enabling adaptive, robust, and physics-aware object representations that open new possibilities for robotic manipulation of complex deformables. Experiments show that EMPM outperforms spring-mass baseline models. Project website: \url{https://embodied-mpm.github.io}.
\end{abstract}

\twocolumn[{%
\renewcommand\twocolumn[1][]{#1}%
\maketitle
\begin{center}
    \centering
    \captionsetup{type=figure}
    \includegraphics[width=\textwidth]{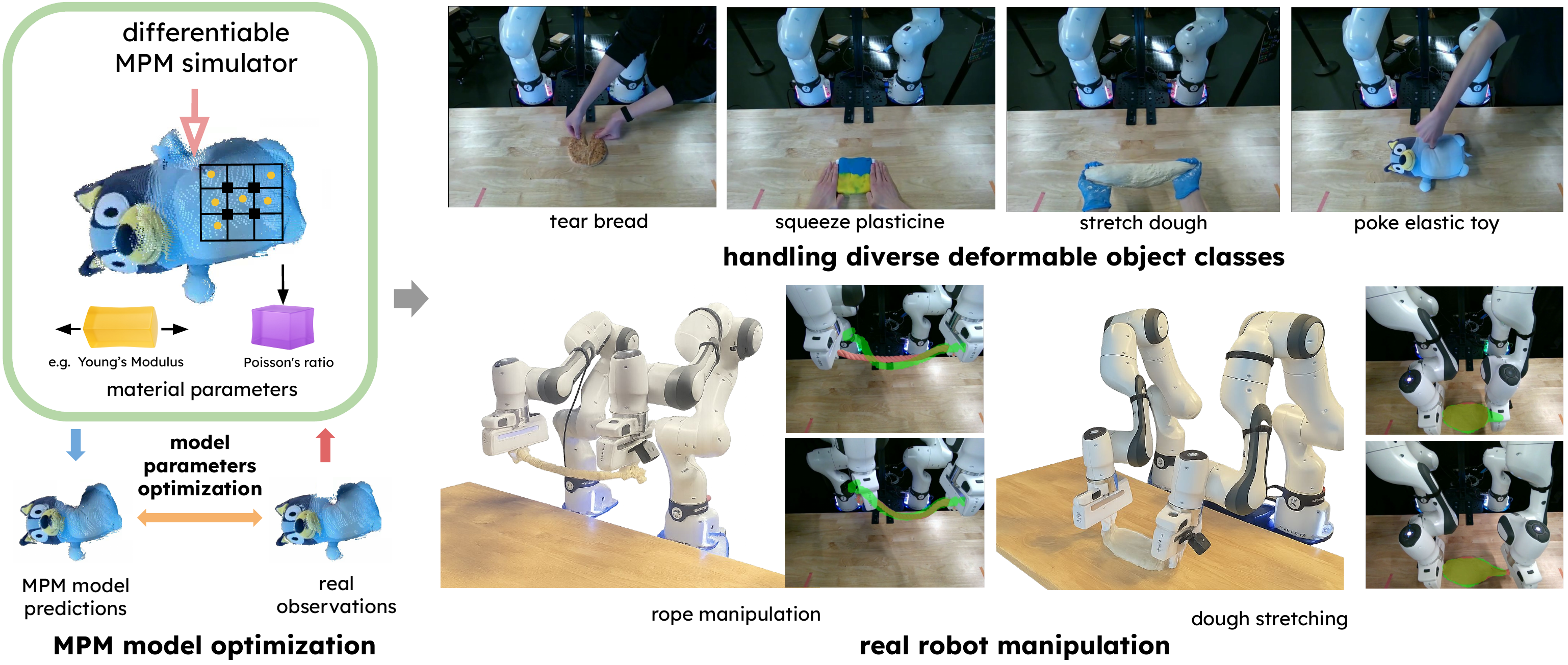}
    \caption{Overview. Our differentiable MPM simulator estimates material parameters from real observations such as 3D object shape. It handles diverse, complex deformable objects and can update parameters online from streaming video. 
    }
    \label{fig:teaser}
\end{center}
}]

\section{Introduction}
Accurately modeling deformable objects, including their geometry, appearance, physical properties, and dynamics from visual information, and using these models for differentiable physics simulation, remains a long-standing and fundamental challenge in 3D vision, graphics, and robotics. By using a differentiable real-to-sim-to-real pipeline, we can employ gradient-based optimization to perform accurate interactive simulation of deformable objects directly from visual inputs, paving the way for constructing photo- and physics-realistic digital twins, as well as enabling physics-aware robotic manipulation tasks.

Joint modeling and simulation of deformable objects often involve integrating geometry reconstruction, appearance rendering, and physics-based simulation. With recent advances in radiance field methods such as NeRF and Gaussian Splatting (GS) \cite{mildenhall2020nerf, kerbl3Dgaussians}, appearance modeling for both static and dynamic scenes and objects has seen significant progress through differentiable rendering techniques \cite{luiten2023dynamic, Matsuki:Murai:etal:CVPR2024}. However, physics-based simulation remains a core challenge in this domain. Previous approaches have tackled this problem using particle-based physics models \cite{Abou-Chakra_2024_WACV, abouchakra-embodiedgaussians, abouchakra2025realissim} and spring-mass systems \cite{jiang2025phystwin}. While effective for certain scenarios, these methods often oversimplify the complex dynamics of deformable objects and struggle to accurately simulate various real-world objects. Although learning-based intuitive physics models \cite{zhang2024dynamics, zhang2024particle, obrist2025pokeflex, huang2025ParticleFormer} offer the potential to capture more complex behaviors through statistical approximation, they typically require large amounts of training data and often generalize poorly to novel objects—limiting their real-world applicability. Recently, more advanced methods based on the Material Point Method (MPM) \cite{jiang2016material} have shown promise in simulating deformable objects composed of continuum materials directly from visual information \cite{xie2023physgaussian, li2023pacnerf, zhang2024physdreamer}.

In this paper, we introduce \PaperName{}, a real-to-sim-to-real pipeline that jointly performs MPM-based differentiable physics modeling and simulation, alongside geometry reconstruction and appearance rendering of deformable objects, including both elastic and elastoplastic materials from multi-view RGB-D video inputs. Using a pre-calibrated multi-view RGB-D camera rig, we reconstruct a 3D point cloud to represent the object's geometry. This point cloud is then used to build a 3D Gaussian Splatting (3DGS) model for high-fidelity appearance rendering.
The resulting geometry-appearance model is integrated with an MPM simulator to capture the object’s dynamic behavior. MPM physical parameters—such as elasticity and stiffness—are optimized by minimizing the discrepancy between real-world observations and simulated rollouts, measured in both 3D geometry and visual appearance throughout the object’s deformation process.
Our approach includes both offline training of the MPM physics model parameters using pre-collected RGB-D data and online correction/adaptation based on live sensory observations. While concurrent work \cite{Yang2025Differentiable} also applies MPM simulation to robotic manipulation, it focuses solely on elastoplastic objects and lacks visual rendering capabilities. In contrast, our method handles a broader range of deformable materials and integrates photorealistic rendering. Experimental results on various deformable objects demonstrate the effectiveness of our MPM-based differentiable physics simulation, outperforming baseline spring-mass models \cite{jiang2025phystwin}.

In summary, the major contributions of this paper are as summarized as follows:
\begin{itemize}
    \item We present a real-to-sim-to-real system based on MPM physics simulation, featuring differentiable parameter optimization grounded in real sensory observations.
    \item We present an online parameter optimization scheme that adapts to incoming observations, unlocking key power for potential downstream robotic control tasks.
    \item We demonstrate our method on a wide range of deformable objects, including both elastic and elastoplastic materials, which are challenging for existing real-to-sim-to-real simulation approaches.
\end{itemize}

\section{Related Work}
\subsection{4D visual reconstruction}
4D reconstruction is the problem of modeling 3D geometry of dynamic objects/scene over time to capture their motion. 
Many methods track dynamic objects in 3D and differentiate the static background from the moving objects. Recent 3D point tracking feedforward models are able to generalize to unseen scenarios and handle certain degrees of occlusion \cite{li2024_MegaSaM, karaev2024cotracker3}. Recent radiance field approaches also adapt this traditional 4D reconstruction framework, extending the original NeRF to dynamics alternatives \cite{pumarola2020d}. Radiance field methods based on explicit scene representation, such as dynamic Gaussian Splatting-based methods, adapt more naturally with the conventional 4D reconstruction more naturally by regularizing the rigid-body motion of underlying Gaussian kernels \cite{luiten2023dynamic}. Our reconstruction pipeline follows these aforementioned methods and consists of point cloud lifting, 3D tracking and GS-based appearance modeling.

\subsection{Physics simulation for deformable objects}
Although physics simulation has been studied for decades \cite{Terzopoulos1987massspring, OBrien1999Graphical, Sifakis2012FEM, jiang2016material}, methods that leverage sensory input and enable differentiability through observations have only gained popularity in recent years \cite{xie2023physgaussian, Feng_2024_CVPR, zhang2024physdreamer, jiang2025phystwin}. Early methods follow particle-based models to model rigid objects and a few types of deformable objects \cite{abouchakra-embodiedgaussians}. Subsequent approaches have adopted the spring–mass model for simulating deformable objects \cite{zhong2024springgaus, jiang2025phystwin}, and learning-based particle/grid-based methods \cite{li2018learning, zhang2024dynamics, zhang2024particle, zhang2024adaptigraph, huang2025ParticleFormer}. As the dynamics of deformable objects and interactive motions such as tearing become more complex, spring–mass and learning-based models may fail to accurately capture the underlying physics. Subsequently, MPM-based approaches emerged as one of the leading approaches in this domain, which are well-suited for handling objects with continuum materials, and interactions like large deformations, fracture and tearing. Differentiable MPM-based methods been successfully demonstrated in prior work \cite{zhang2024physdreamer}, which likewise underpins our work. 

\subsection{Differentiable MPM simulation}
To accurately model and simulate real-world deformable objects, it is often necessary to optimize physical parameters such as material stiffness based on real-world observations, a process commonly referred to as system identification. The sources of discrepancy used to guide this optimization can vary, including force measurements \cite{xu2023differentiable}, 3D geometry alignment \cite{jiang2025phystwin}, and photometric loss from image-based rendering \cite{li2023pacnerf, zhang2024physdreamer}. Our method follows this line of research by leveraging 3D geometry to optimize MPM parameters in a differentiable manner. 
While deriving analytical gradients for multi-stage physical simulators is prohibitively complex, recent advances have shown that auto-differentiation can effectively make MPM simulation a differentiable layer \cite{hu2019chainqueen}. Most recent efforts in applying differentiable MPM systems have leveraged frameworks such as Taichi \cite{hu2019taichi, hu2019difftaichi} and Warp \cite{warp2022}, demonstrating reliable capabilities in physical parameter estimation and the advantages of GPU acceleration \cite{li2023pacnerf, zhang2024physdreamer, su2023generalized, zong2023neural}. \rv{Similarly, we also leverage Warp \cite{warp2022} as the tool for our differentiable MPM engine to differentiate through the model outputs.}

\section{Method}

\begin{figure*}[ht!]
    \centering
    \includegraphics[width=\linewidth]{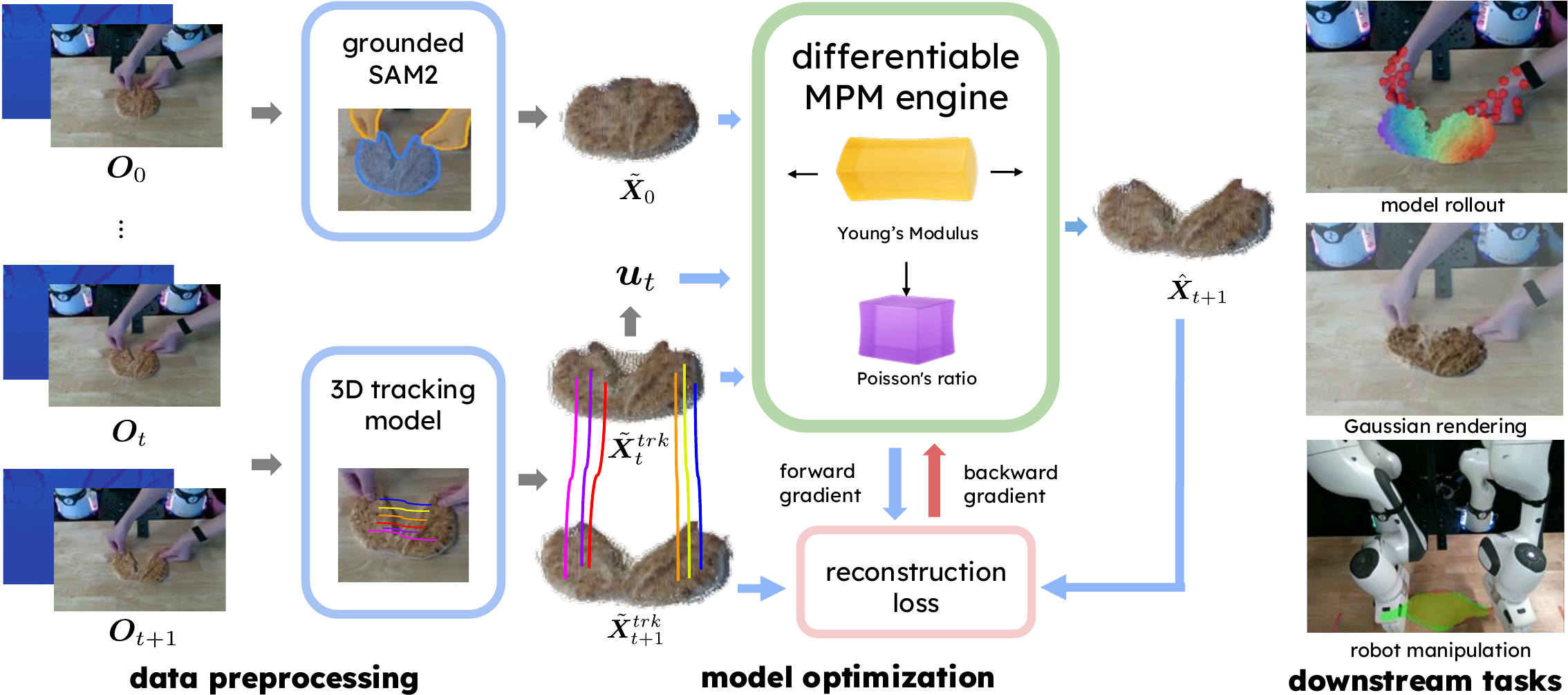}
    \caption{Method Overview of \PaperName{}. Our MPM simulation engine takes a reconstructed pointcloud and tracked 3D points as inputs. The model parameters are optimized using the discrepancy between the model's prediction and 3D shape reconstruction and tracking. The control inputs applied to the dynamics model of MPM is computed as the velocity extracted from the reconstructions. In the inference stage, we test the model's performance in terms of predicting 3D point positions and RGB image rendering of the 3D Gaussian splats.}
    \label{fig:method_overview}
\end{figure*}

\subsection{Overview}
We model deformable object dynamics with an action-conditioned MPM simulator that is \emph{embodied} through robot (or hand) interactions. Let $\bm{X}_t$ denote particle states at time $t$ (with positions $\bm{x}_p$, velocities $\bm{v}_p$, and deformation gradients $\bm{F}_p$), and let $\bm{u}_t$ denote controller (hands or grippers) actions obtained from hand tracking or robot telemetry. The MPM model advances state prediction at the next step as,
$
\hat{\bm{X}}_{t+1} \;=\; f_{\boldsymbol{\theta}}\!\left(\bm{X}_{t},\, \bm{u}_{t}\right)
$, 
where $\boldsymbol{\theta}$ denotes material parameters (e.g., Young’s modulus $\bm{E}$, Poisson’s ratio $\bm{\nu}$, density $\bm{\rho}$, plastic yield stress $\bm{y}$). 

We use multi-view RGB‐D images to provide observations $\bm{O}^i_t = \{\bm{I}^i_t,\bm{D}^i_t\}$, where $\bm{I}$ and $\bm{D}$ denote color and depth images, and $i$ denotes the camera index. We reconstruct the point cloud $\tilde{\bm{X}}_t \in \mathbb{R}^{3 \times N}$ by fusing the points and uplifting them to 3D world frame. Due to potential occlusions, we employ a point tracking model using the initial frame as a reference to obtain the 3D point trajectories $\tilde{\bm{X}}^{\text{trk}}_t$.

Our goal is to optimize the MPM model parameters $\boldsymbol{\theta}$, based on the loss function of observed points $\tilde{\bm{X}}$, $\tilde{\bm{X}}^{\text{trk}}$ and model predictions $\hat{\bm{X}}$, hence solving the general optimization problem in the following form:
\begin{equation}
    \operatorname*{argmin}_{\boldsymbol{\theta}} L \left( \tilde{\bm{X}}, \tilde{\bm{X}}^{\text{trk}}, f_{\boldsymbol{\theta}} (\bm{X}, \bm{u}) \right)\label{eq:optimization}
\end{equation}

\subsection{MPM and Continuum Mechanics}
We briefly review the Material Point Method (MPM)~\cite{jiang2016material} used in our work. MPM couples continuum balance laws with a hybrid particle–grid discretization, making it well-suited for large deformations, contact, and elastoplastic flow in embodied interactions.

MPM represents objects as particles while performing force computations on a background grid. At each step, particle mass and momentum are transferred to grid nodes (P2G)\rv{, in eq. \eqref{eq:p2g}}. Grid velocities are then updated with action conditions\rv{(eq. \eqref{eq:grid_update})} and transferred back to the particles for advection (G2P)\rv{, in eq. \eqref{eq:g2p}}. We summarize the explicit MPM update at time step $t^n$ as follows, with $p$ denoting particles and $i$ denoting grid nodes:

\noindent\textbf{P2G (particle to grid).}
We transfer mass and momentum using the affine particle-in-cell method (APIC) \cite{jiang2015affine} scheme:
\begin{equation}
\small
\begin{aligned}
m_i^n = \sum_p w_{ip}^n\, m_p, \hspace{2mm}
m_i^n \bm{v}_i^n = \sum_p w_{ip}^n\, m_p \Big(\bm{v}_p^n + \bm{C}_p^n(\bm{x}_i-\bm{x}_p^n)\Big),
\end{aligned}
\label{eq:p2g}
\end{equation}
where $w_{ip}$ are the B-spline weights and $\bm{C}_p^n$ is the affine velocity term.

\noindent\textbf{Grid update (embodiment and contact).}
Grid velocities advance under stress, gravity, and constraints:
\begin{equation}
\bm{v}_i^{n+1} \;=\; \bm{v}_i^n + \frac{\Delta t}{m_i}\,\bm{f}_i(\bm{x}_i^n;\boldsymbol{\theta}).
\label{eq:grid_update}
\end{equation}
\rv{Here, \(\bm{f}_i\) denotes the forces at grid node \(i\), which are internally computed within \(f_{\boldsymbol{\theta}}\), the full simulation step in eq.~\eqref{eq:optimization}.}
We enforce \emph{Dirichlet boundary velocities} on nodes in contact with the controller (from reconstructed or robot-measured end-effector motion $\bm{u}_t$), and apply Coulomb friction for table/gripper contact.

\noindent\textbf{G2P (grid to particle).} Grid velocities are transferred back to particles:
\begin{equation}
\bm{v}_p^{n+1} = \sum_i w_{ip}^n \bm{v}_i^{n+1}, 
\qquad
\bm{x}_p^{n+1} = \bm{x}_p^{n} + \Delta t\,\bm{v}_p^{n+1}.
\label{eq:g2p}
\end{equation}

\noindent\textbf{Deformation update.} Deformation gradient is updated by:
\begin{equation}
\bm{F}_p^{n+1} \;=\; \Big(\bm{I} + \Delta t \sum_i \bm{v}_i^{n+1} (\nabla w_{ip}^n)^\top \Big)\bm{F}_p^{n}.
\end{equation}

\noindent\textbf{Elastoplasticity.}
Most real-world deformable objects are not purely elastic and they undergo permanent shape changes under deformation. In continuum mechanics, this can be modeled by factoring the deformation gradient as \(\bm{F} = \bm{F}^E \bm{F}^P\). Here the plastic part \(\bm{F}^p\) serves as an intermediate state that stores the new rest shape, and only the elastic part \(\bm{F}^E\) is used to compute stress. During MPM updates, we use a trial deformation graident to track this process:
\begin{equation}
\footnotesize
\begin{aligned}
& \bm{F}_p^{\text{trial}, n+1} = \text{updateF}(\bm{F}_p^{E, n}),
& \bm{F}_p^{E, n+1} = \text{returnMap}(\bm{F}_p^{\text{trial}, n+1})
\end{aligned}
\end{equation}
In this paper, we use Fixed Corotated stress as elasticity model and use von Mises return mapping to model plasticity.

\noindent\textbf{Connection to embodiments.}
Robot or hand motion $\bm{u}_t$ can enter as boundary velocities at the grid step. This makes the simulator \emph{action-conditioned}: given candidate end-effector trajectories $\bm{u}_{t:t+H}$, we can predict object responses $\hat{\bm{X}}_{t:t+H}$ and score task costs (target geometry, collision). The same interface supports system identification (align to observations) and decision-making (rollout for planning). We next instantiate this in the offline identification setting.

\subsection{Offline Simulation and Optimization}
We instantiate the action-conditioned simulator for parameter identification from multi-view RGB-D videos. At time step $t$, RGB-D image observations $\{\bm{I}^i_t,\bm{D}^i_t\}_{t=0}^{T}$ are fused into point sets $\tilde{\bm{X}}_t$, and end-effector motion is reconstructed as $\bm{u}_{0:T-1}$, which is enforced as boundary velocities in the grid update.

\noindent\textbf{Problem.}
Let $\bm{X}_0$ denote particles sampled from $\tilde{\bm{X}}_0$ (with point densification applied when required for enhanced dynamics). For material parameters
\(
\boldsymbol{\theta}=\{\bm{E}, \bm{\nu}, \bm{\rho}, \bm{y} \},
\)
the MPM rollout recursively produces $\hat{\bm{X}}_{1:T} = f_{\boldsymbol{\theta}}(\bm{X}_0,\bm{u}_{0:T-1})$. The identification problem is
\begin{equation}
\boldsymbol{\theta}^{\star}
\;=\;
\operatorname*{arg\,min}_{\boldsymbol{\theta}}
\; L_{\text{offline}}\!\big(\tilde{\bm{X}}_{1:T},\,\tilde{\bm{X}}^{\text{trk}}_{1:T},\, \hat{\bm{X}}_{1:T}\big),
\end{equation}
which conditions learning on the observed interaction through $\bm{u}_{0:T-1}$.

\noindent\textbf{Real-to-sim data processing.}
We use grounded-SAM2~\cite{ravi2024sam2segmentimages} to compute the object and controllers (e.g. hands) masks for each image frame based on language prompts. Next we lift the 3D object point cloud measurement $\tilde{\bm{X}}_t$ with depth and camera extrinsics. 3D temporal correspondences $\tilde{\bm{X}}^{\text{trk}}_t$ are computed on offline sequences with an off-the-shelf point tracking model. The controller actions $\bm{u}_t$ are bound to grid nodes as Dirichlet boundary velocities, and contacts with the table and grippers are modeled using Coulomb friction.

\noindent\textbf{Objective.}
To bridge the discrepancy between MPM model predictions and real observations, we minimize the predicted position $\hat{\bm{X}}_t$ with lifted 3D points and their trackings:
\begin{equation}
\footnotesize
\begin{aligned}
L_{\text{offline}} &=
\lambda_{\text{dist}} \sum_{t=1}^{T} \mathrm{Chamfer}\!\left(\hat{\bm{X}}_t, \tilde{\bm{X}}_t\right) + \lambda_{\text{trk}} \sum_{t=1}^{T}\sum_{j\in\mathcal{T}_t}
\left\|\hat{\bm{X}}_{t,j}-\tilde{\bm{X}}^{\text{trk}}_{t,j}\right\|_2^2 
\end{aligned}
\end{equation}
In this loss function, the first term measures the 3D Chamfer distance between the point clouds, while the second term computes the \(\mathcal{L}^2\) squared distance between each particle, serving to align the motions of individual particles.
We use occlusion masks from the pre-processing stage to filter out the invalid particles and exclude them from loss computation.

\noindent\textbf{Optimization.}
We differentiate through the model rollout with Warp~\cite{warp2022} to obtain the loss gradient $\nabla_{\boldsymbol{\theta}}L_{\text{offline}}$ and update $\boldsymbol{\theta}$ by gradient-based optimization methods. While gradient-based optimization enables smooth recovery of MPM physics parameters, we also provide an option of zero-order optimization using CMA-ES on the forward simulator. This strategy may be preferable for reducing memory usage or enabling a more aggressive parameter search. 

\subsection{Online Adaptive Identification}
We also optimize the model parameters online during real interactions using streaming 3D observations.

\noindent\textbf{Streaming supervision from cameras.}
At the beginning of streaming, we use multi-view RGB-D images to fuse an initial point cloud \(\tilde{\bm{X}}_0\), which is then sampled as the MPM particles \({\bm{X}}\). Similar to offline optimization, we use Grounded SAM2 model to segment target objects. In subsequent streaming steps, we continuously perform segmentation and back-projection to lift the video frames to get 3D point clouds, serving as the 3D observation \(\tilde{\bm{X}}_t\). Additionally, we record the current positions of the robot grippers as the control inputs \(\bm{u}_t\) to define the boundary conditions.
During online streaming, we observe that 3D point tracking becomes less reliable as tracked points tend to become invalid over time due to occlusion in indefinite incoming image sequences. Therefore, we do not perform online key point tracking in this stage.

\noindent\textbf{Simulation and Online Correction.}
With the object geometry reconstructed in real time, we perform simultaneous simulation on the initial point cloud \(\tilde{\bm{X}}_0\) and the current controller motion \(\bm{u}_t\). Through continuous dynamic updates, the simulated object \(\hat{\bm{X}}_t\) serves as a synchronized digital twin, closely aligning with the real object's motion.

To narrow the gap in real-to-sim object deformation and extract estimation of material parameters of the modeled objects, we perform differentiable-simulation steps along with the synchronized simulation, generating gradient flows to direct the object to 3D observation. To avoid unstable gradients, we restrict the optimizing steps to only \textit{quasi-static} states -- assuming objects are close to stable-equilibrium. \rv{More specifically, when an object is under balanced forces, such as being pinned by two end-effectors, we perform a fixed-step \(H\) forward simulation, using the current particle states as input: 
\(\check{\bm{X}}_{t+H} = f_{\bm{\theta}}(\hat{\bm{X}}_t, \bm{u}_t).\) Under the equilibrium assumption, the object should not deviate from its current rest shape (i.e., \(\check{\bm{X}}_{t+H}\approx \hat{\bm{X}}_t\)) if the material parameters are estimated correctly. To penalize the discrepancy caused by inaccurate parameters, we propose a loss function similar to the one used in offline optimization. We use the same 3D Chamfer distance \(L_{\text{dist}}\) to align the point clouds. Since we do not have direct correspondence between simulated and observed particles during online optimization, we additionally employ a 2D segmentation mask loss \(L_{\text{mask}}\) as a regularizer to align the simulated particles \(\check{\bm{X}}_{t+H}\) with the segmented mask \(\bm{M}_{t+H}\). Together, we have:}
\begin{equation}
L_{\text{online}} = \lambda_{\text{dist}} L_{\text{dist}} + \lambda_{\text{mask}} L_{\text{mask}}.
\label{eq:online_loss}
\end{equation}
Similarly, we back-propagate the gradient to obtain a new estimation of parameter \(\bm{\theta}'\). After each step, the new parameter \(\bm{\theta}'\) will replace the current parameter in the simulation.

\subsection{Applications}
The \PaperName{} simulator, potentially combined with offline and online optimization, offers a direct interface for incorporating deformable dynamics into embodied workflows. Because end-effector motions are bound to grid boundary conditions, \PaperName{} produces action-conditioned rollouts that can potential be queried for both interactive simulation and robotic decision making.

With the digital twin synchronized to observed motions, operators can visualize the responses of objects under teleoperation or hypothetical controller inputs. This enables rapid what-if analysis of manipulation strategies and immediate feedback on feasibility, contact regimes, and deformation patterns.
Candidate end-effector trajectories $\{\bm{u}_{t:t+H}\}$ are applied as boundary velocities to generate predictions $\hat{\bm{X}}_{t:t+H}$. 
The same interface could support both gradient-free sampling and trajectory optimization.
End-effector control points and simulator boundary conditions are aligned in \PaperName{}, which transforms the simulator into a predictive engine that seamlessly connects perception to planning and execution.

\section{Experiments}

\subsection{Experiment Setup}
\begin{figure}[ht!]
    \centering
    \includegraphics[width=0.95\linewidth]{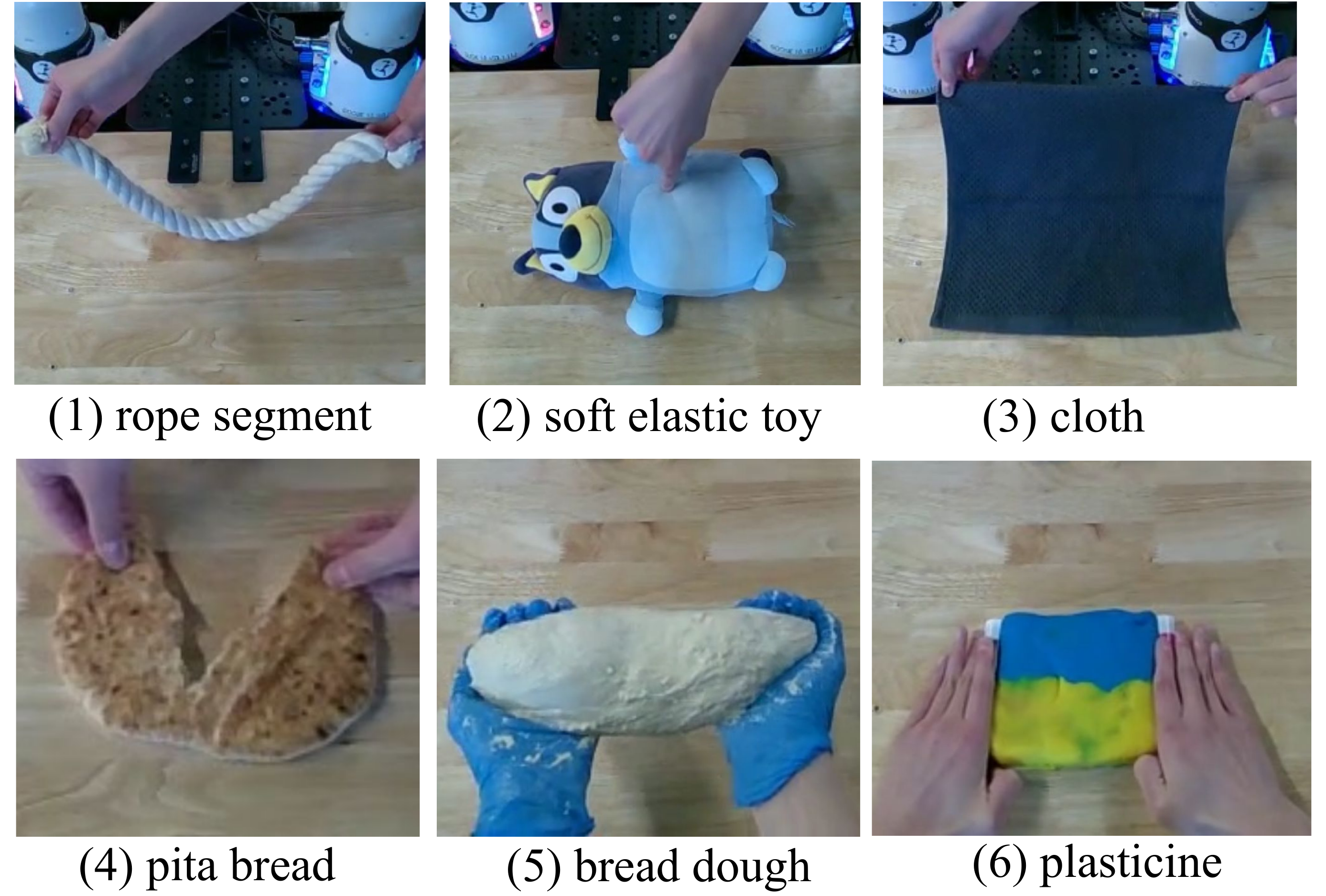}
    \caption{Objects used in our experiments. first row: elastic objects to simulate bending, poking and pulling; second row: elastoplastic objects to simulate fracture, stretch, and squeezing.}
    \label{fig:simulated_objects}
\end{figure}

To demonstrate the coverage and capability of the MPM simulation engine in modeling deformable objects, we select objects from two major categories:
\begin{itemize}
\item \textbf{Elastic} objects: a rope segment; a piece of cloth; a soft elastic toy that bounces back when pushed or poked
\item \textbf{Elastoplastic} objects: a piece of plasticine; a lump of bread dough; a pita bread
\end{itemize}
Figure~\ref{fig:simulated_objects} shows the types of objects we simulate in our experiments. We perform various manipulations on these objects—including extending, stretching, squeezing, and tearing—to demonstrate the range of deformations achievable in our framework.

For the sensor setup used in data collection and online optimization, we employ a multi-view stereo RGB-D camera rig as the perception system. In our experiments, we set up three RealSense D455 cameras. We first collect data using human hands to facilitate the complicated interactions with the deformable objects that robot grippers struggle. This data is used in the offline simulation and optimization pipeline. For the robotic manipulation experiments within the online optimization framework, we use a bimanual setup with two Franka arms equipped with their original grippers.

\rv{\textit{Implementation details}: We perform training/online optimizaiton of the MPM model parameters on a single Nvidia A6000 GPU. We optimize the Young's modulus \(E\) and Poisson's ratio \(\nu\) of the objects and assume the parameters remain constant accross the material field. \(E\) is normalized to \([0,1]\) by a factor of \(10^6\). We leverage Warp’s automatic differentiation capability via its PyTorch integration, using the AdamW optimizer with a default learning rate of \(10^{-4}\).}

\subsection{Offline Simulation and Optimization}
Using the collected videos, we follow the data processing procedure outlined above to extract the 3D information and perform offline simulation and optimization.

\begin{figure*}[ht!]
    \centering
    \includegraphics[width=\linewidth]{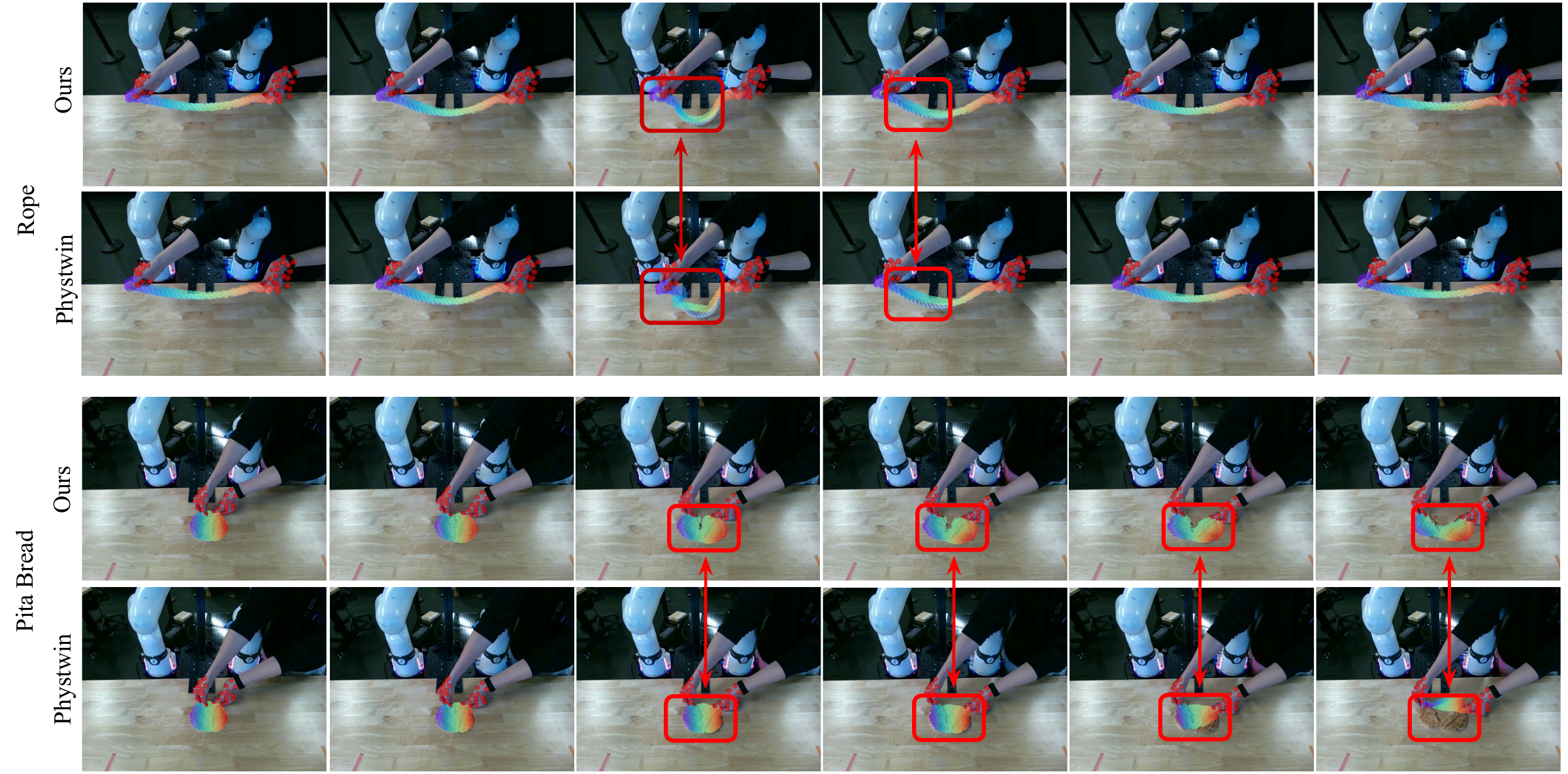}
    \caption{Qualitative results showing the simulation predictions and real observations for different objects, comparing our method and PhysTwin \cite{jiang2025phystwin}. Our method especially shines at modeling fracture, illustrated in the pita bread example, where PhysTwin model prediction completely fails. In the last frame of the plasticine example, the PhysTwin model prediction fails due to the over-bend of the internal spring model, where our model can still successfully model the squeezing of the object.}
    \label{fig:quali_offline}
\end{figure*}

\begin{figure*}[ht]
    \centering
    \includegraphics[width=\linewidth]{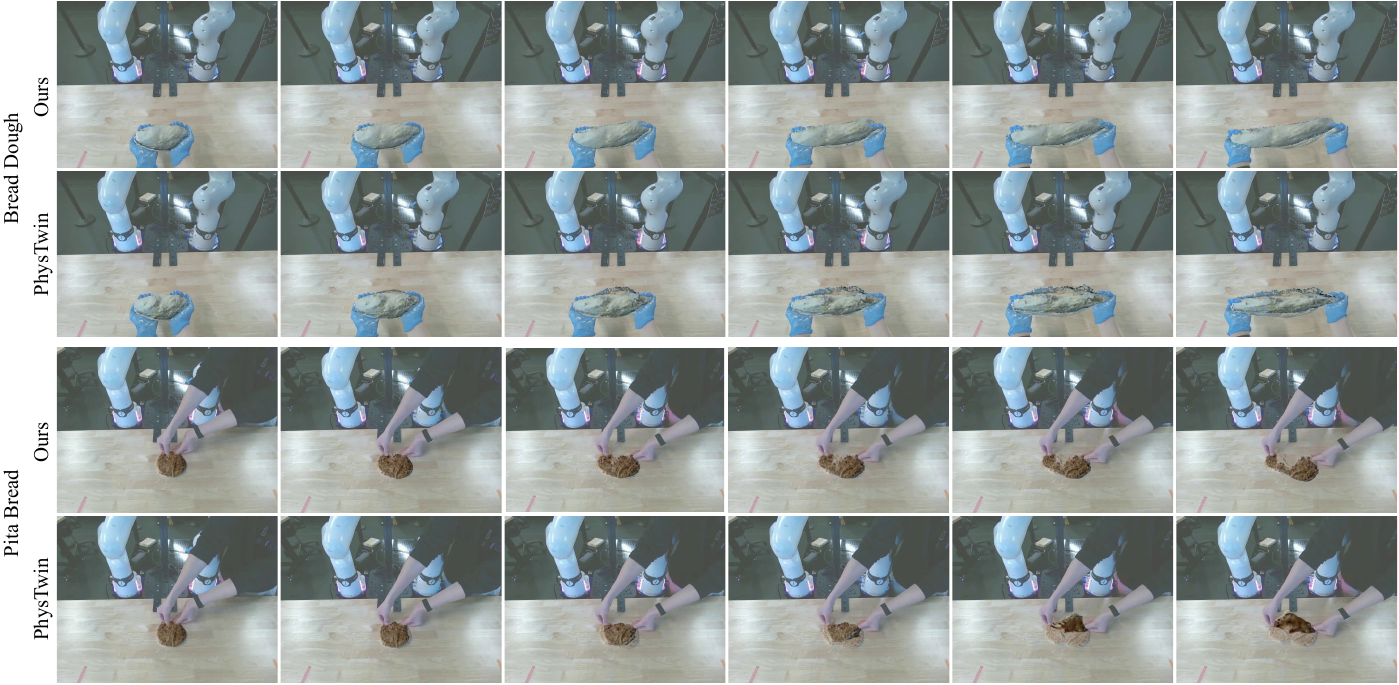}
    \caption{Qualitative results demonstrate photorealistic reconstruction using Gaussian Splatting. Our reconstructed dynamics closely match real observations, whereas the spring-mass-based PhysTwin\cite{jiang2025phystwin} fails to accurately model these materials.}
    \label{fig:quali_gaussian}
\end{figure*}

We visualize the simulated particles overlaid on the input video in Figure~\ref{fig:quali_offline}. Our simulated results exhibit physically plausible deformations consistent with the properties of each object type, closely aligning with the behavior observed in the real world. Additionally, we present Gaussian-rendered visualizations of the dynamic simulations in Figure~\ref{fig:quali_gaussian}, offering a photorealistic reproduction of the visual observations. \rv{The Gaussian splatting is based on the origianl 3DGS and the gsplat \cite{ye2024gsplat} library's implementations and the motion of the Gaussian centers is interpolated with the motions of neighboring
MPM particle nodes using Linear Blend Skinning (LBS) \cite{jiang2025phystwin}.}

To demonstrate the advantages of our method, we compare our results with PhysTwin\cite{jiang2025phystwin}, which models deformable objects using a spring–mass system. We visualize the qualitative results and photorealistic rendering in Figures~\ref{fig:quali_offline} and \ref{fig:quali_gaussian}. To quantitatively evaluate the performance, we report the reconstruction qualities in terms of geometric differences of point clouds, as well as visual rendering qualities. We also compare our results with a learning-based dynamic model PGND\cite{zhang2024particle}, reported in Table~\ref{tab:comparison}.

Leveraging our MPM-based simulator, our method achieves higher simulation accuracy and narrows the gap between real-world captured data and the simulated prediction. Compared to the spring–mass system baseline \cite{jiang2025phystwin} and neural dynamic baseline \cite{zhang2024particle}, our approach demonstrates better alignment in both geometric and visual metrics.

In addition to offering higher overall simulation accuracy, another key advantage of EMPM is its ability to handle a broad spectrum of materials and physical phenomena within our framework. Our method can model both hyperelastic and elastoplastic materials, encompassing a wide range of everyday objects—from purely elastic jelly to plasticine, which exhibits largely different behavior under external forces. In contrast, spring-mass systems are limited to modeling elastic springs and cannot capture intrinsic non-elastic behaviors, such as permanent shape change beyond the elastic threshold in elastoplastic materials, as demonstrated in our experiments. Moreover, spring-mass systems are unable to simulate more complex phenomena like material damage (e.g., in the pita experiment), since elastic springs do not naturally accommodate fracture. On the other hand, our method accurately captures fracture dynamics and produces realistic and desirable outcomes.

\begin{table*}[ht!]
\centering
\begin{tabular}{lcccccccccccc}
\toprule
\multirow{2}{*}{\textbf{Method}} & \multicolumn{6}{c}{\textbf{Elastic Objects}} & \multicolumn{6}{c}{\textbf{Elastoplasic Objects}} \\
\cmidrule(lr){2-7} \cmidrule(lr){8-13}
& Dist \(\downarrow\) & Track\(\downarrow\) & IoU \% $\uparrow$ & PSNR $\uparrow$ & SSIM $\uparrow$ & LPIPS \(\downarrow\) 
  & Dist \(\downarrow\) & Track\(\downarrow\) & IoU \% $\uparrow$ & PSNR $\uparrow$ & SSIM $\uparrow$ & LPIPS \(\downarrow\) \\
\midrule
PGND \cite{zhang2024particle} & 0.0618 & - & 0.6898 & 21.06 & 0.8938 & 0.0977 & 0.0245 & - & 0.5069 & 21.26 & 0.9472 & 0.0738 \\

PhysTwin \cite{jiang2025phystwin} & 0.0227 & 0.1467 & 0.6981 & 24.12 & 0.9312 & 0.0756 & 0.0177 & 0.1108 & 0.6918 & 27.01 & 0.9695 & 0.0353 \\

Ours & \textbf{0.0222} & \textbf{0.1377} & \textbf{0.7095} & \textbf{24.19} & \textbf{0.9319} & \textbf{0.0711} & \textbf{0.0082} & \textbf{0.1014} & \textbf{0.7768} & \textbf{27.82} & \textbf{0.9725} & \textbf{0.0291} \\
\bottomrule
\end{tabular}
\caption{Quantitative Comparison on Offline Simulation. We compare our prediction results with PhysTwin\cite{jiang2025phystwin} and PGND\cite{zhang2024particle} in terms of both reconstruction accuracy (measured by distance and tracking of the 3D point cloud; tracking errors from \cite{zhang2024particle} are omitted because we do not have point correspondence between their results and ours) and visual quality (evaluated using IoU, PSNR, SSIM, and LPIPS) through photorealistic rendering with Gaussian Splatting. The results are categorized by object type: elastic and elastoplastic. Our method outperforms baselines across all metrics, with a particularly notable advantage in handling elastoplastic objects.}
\label{tab:comparison}
\end{table*}

\subsection{Online Optimization}
\begin{figure*}[ht]
    \centering
    \includegraphics[width=\linewidth]{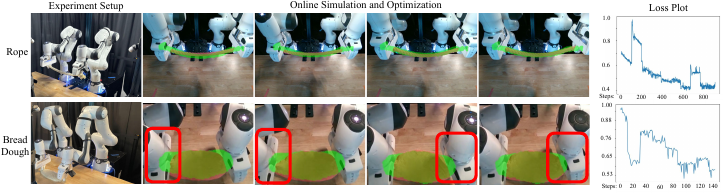}
    \caption{Qualitative results of the online simulation demonstrate the tracking performance of our MPM model against real observations. With online optimization guided by real-time reconstruction, the alignment between simulation and observation improves progressively over time. The losses are normalized from 0 to 1 for better illustration. Updates in the bread dough case are not obvious due to a relatively small motion and the occlusion of the grippers, it is recommended to refer to the video for a better dynamic visualization.}
    \label{fig:quali_online}
\end{figure*}

To demonstrate the efficacy of our online optimization, we conduct experiments on one elastic object (rope) and one elastoplastic object (bread dough). We teleoperate a pair of bimanual Franka arms to control the objects. Specifically, we control the arms to move along various trajectories, while also allowing the objects to remain steady between motions. With an initial point cloud generated at the beginning of the loop, we simulate the object point clouds under the same manipulations. As shown in Figure~\ref{fig:quali_online}, our simulated twins closely follow the motion of the real objects.

We perform online differentiable simulation to optimize the material stiffness of the objects during the simulation, with the loss function shown in Eq. \ref{eq:online_loss}. In our experiments, we perform one optimization step every 5 online streaming step and provided the object is relatively steady at that step. In each optimization step, we simulate for a total of 10 forward simulation steps. With the optimization, the alignments between simulated objects and real observations improved over time, while we gradually adjust our estimation of the physical parameters.
To quantitatively evaluate the improvement with simulation correction, we report the 2D mask and 3D distance errors in the online experiments for both objects, with and without optimization, in Table~\ref{tab:online}. We observe improved simulation accuracy with the presence of our online correction.

\begin{table}[ht]
\vspace{-0.5em}
\centering
\begin{tabular}{lcccc}
\toprule
\multirow{2}{*}{\textbf{Experiment}} & \multicolumn{2}{c}{\textbf{Without Optimization}} & \multicolumn{2}{c}{\textbf{With Optimization}} \\
\cmidrule(lr){2-3} \cmidrule(lr){4-5}
& \(L_{\text{mask}}\) \(\downarrow\) & \(L_{\text{dist}}\) \(\downarrow\) 
  & \(L_{\text{mask}}\) \(\downarrow\) & \(L_{\text{dist}}\) \(\downarrow\)  \\
\midrule
Rope & 0.0456 & 0.0057 & \textbf{0.0428} & \textbf{0.0054}\\
Bread Dough & 0.0031 & 0.0060 & \textbf{0.0024} & \textbf{0.0059} \\

\bottomrule
\end{tabular}
\caption{Quantitative results demonstrate improvements in geometric accuracy achieved through progressive correction during online simulation.}
\label{tab:online}
\end{table}

\rv{We also include running time analysis results comparing \PaperName{} with the baseline methods PhysTwin \cite{jiang2025phystwin} and PGND \cite{zhang2024particle} in Table~\ref{tab:running_time}. For PhysTwin, we conducted both CMA-ES and gradient-based optimization and report only the gradient training time. For PGND, we train 10 episodes following the default setup. The training of \PaperName{} is the most efficient due to its optimization over entire training frames as a batch. PGND achieves the lowest inference time due to the feedforward pass of a relatively small network, but with the cost of very long training time.}
\begin{table}[ht]
\vspace{-10pt}
\centering
\rv{%
\begin{tabular}{lcccc}
\toprule
\multirow{2}{*}{\textbf{Running time (s)}} & \multicolumn{2}{c}{\textbf{Elastic Objects}} & \multicolumn{2}{c}{\textbf{Elastoplasic Objects}} \\
\cmidrule(lr){2-3} \cmidrule(lr){4-5}
& Training \(\downarrow\) & Testing \(\downarrow\)
  & Training \(\downarrow\) & Testing \(\downarrow\)  \\
\midrule
PGND \cite{zhang2024particle} &50675.50  &\textbf{5.06} &48464.40 &\textbf{5.34} \\
PhysTwin \cite{jiang2025phystwin} &496.84  &11.49 &589.51 &15.94  \\
Ours &\textbf{161.80} &14.71 &\textbf{171.60} &22.21 \\
\bottomrule
\end{tabular}%
}
\caption{\rv{Running time (in seconds) analysis of \PaperName{} and baseline approaches using the same trianing data.}}
\label{tab:running_time}
\end{table}

\subsection{Applications}
\begin{figure}[ht]
    \centering
    \centering
    \begin{subfigure}{\linewidth}
        \centering
        \includegraphics[width=\linewidth]{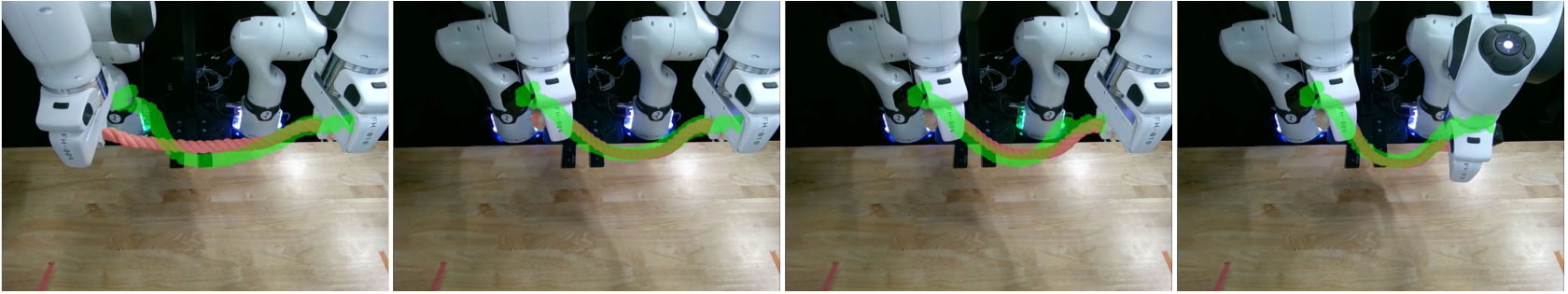}
        \caption{Rope operation}
        \label{fig:sub1}
    \end{subfigure}
        
    \begin{subfigure}{\linewidth}
        \centering
        \includegraphics[width=\linewidth]{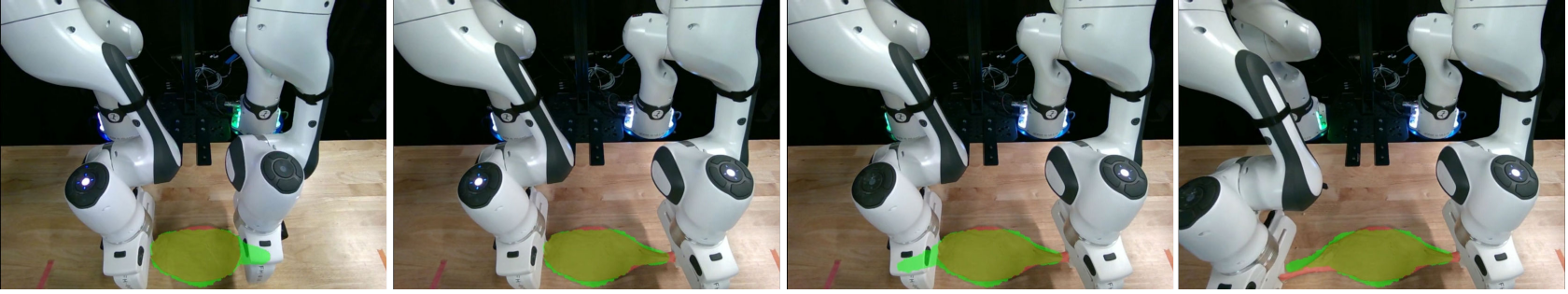}
        \caption{Bread dough operation}
        \label{fig:sub2}
    \end{subfigure}
    
    \caption{Application of \PaperName{} in potential model-based planning and control for manipulation. Here we simulate model rollout and let robot track these changes.}
    \label{fig:apps}
\end{figure}

In our experiment, we demonstrate a proof-of-concept application with real robotics hardware to show how our method would be used for deformable object manipulation tasks, shown in Figure \ref{fig:apps}. With accurate dynamics model for deformable objects proposed in this work, we can potentially deploy model-based robotic motion planning and control and enable autonomous manipulations.

\section{Conclusions \& Discussions}
We presented \PaperName{}, an embodied, differentiable MPM framework for modeling geometry, appearance, and dynamics of complex deformable objects -- including elastic and soft/hard elastoplastic materials that are difficult for spring–mass baselines. EMPM updates material parameters from real observations in both offline and online settings and improves over spring-mass baselines qualitatively and quantitatively. Future work includes enhanced appearance modeling for Gaussian Splatting and integrating model-based control for autonomous deformable manipulation.
\rv{
\paragraph*{Limitation}
We observe that a key obstacle during optimization is obtaining reliable point tracking, whose quality degrades under occlusion and large deformations, leading to unreliable 3D trajectories. Tracking conditioned on the first frame often yields zero reliably tracked points after just a few seconds, rendering it unsuitable for online optimization. We note that emerging tracking models hold promise for mitigating this issue, potentially enabling more faithful 3D guidance for reconstruction and facilitating the integration of tracking-based losses into online optimization.
}

{\small
\bibliographystyle{unsrt} 
\bibliography{main}

@inproceedings{mildenhall2020nerf,
 title={NeRF: Representing Scenes as Neural Radiance Fields for View Synthesis},
 author={Ben Mildenhall and Pratul P. Srinivasan and Matthew Tancik and Jonathan T. Barron and Ravi Ramamoorthi and Ren Ng},
 year={2020},
 booktitle={ECCV},
}

@Article{kerbl3Dgaussians,
      author       = {Kerbl, Bernhard and Kopanas, Georgios and Leimk{\"u}hler, Thomas and Drettakis, George},
      title        = {3D Gaussian Splatting for Real-Time Radiance Field Rendering},
      journal      = {ACM Transactions on Graphics},
      number       = {4},
      volume       = {42},
      month        = {July},
      year         = {2023},
      url          = {https://repo-sam.inria.fr/fungraph/3d-gaussian-splatting/}
}

@inproceedings{luiten2023dynamic,
  title={Dynamic 3D Gaussians: Tracking by Persistent Dynamic View Synthesis},
  author={Luiten, Jonathon and Kopanas, Georgios and Leibe, Bastian and Ramanan, Deva},
  booktitle={3DV},
  year={2024}
}

@inproceedings{Matsuki:Murai:etal:CVPR2024,
  title={{G}aussian {S}platting {SLAM}},
  author={Hidenobu Matsuki and Riku Murai and Paul H. J. Kelly and Andrew J. Davison},
  booktitle={Proceedings of the IEEE/CVF Conference on Computer Vision and Pattern Recognition},
  year={2024}
}

@inproceedings{pumarola2020d,
  title={{D-NeRF: Neural Radiance Fields for Dynamic Scenes}},
  author={Pumarola, Albert and Corona, Enric and Pons-Moll, Gerard and Moreno-Noguer, Francesc},
  booktitle={Proceedings of the IEEE/CVF Conference on Computer Vision and Pattern Recognition},
  year={2021}
}

@inproceedings{Terzopoulos1987massspring,
      title={Elastically deformable models}, 
      author={Demetri Terzopoulos and John Platt and Alan Barr and Kurt Fleischer},
      journal={SIGGRAPH},
      year={1987},
}

@inproceedings{OBrien1999Graphical,
      title={Graphical Modeling and Animation of Brittle Fracture}, 
      author={James F. O’Brien Jessica K. Hodgins},
      journal={SIGGRAPH},
      year={1999},
}

@inproceedings{Sifakis2012FEM,
      title={FEM Simulation of 3D Deformable Solids: A practitioner's guide to theory, discretization and model reduction}, 
      author={Eftychios Sifakis and Jernej BarbicAuthors},
      journal={SIGGRAPH 2012 Courses},
      year={2012},
}

@inproceedings{li2018learning,
    Title={Learning Particle Dynamics for Manipulating Rigid Bodies, Deformable Objects, and Fluids},
    Author={Li, Yunzhu and Wu, Jiajun and Tedrake, Russ and Tenenbaum, Joshua B and Torralba, Antonio},
    Booktitle = {ICLR},
    Year = {2019}
}

@inproceedings{li2023pacnerf,
      title={PAC-NeRF: Physics Augmented Continuum Neural Radiance Fields for Geometry-Agnostic System Identification}, 
      author={Xuan Li and Yi-Ling Qiao and Peter Yichen Chen and Krishna Murthy Jatavallabhula and Ming Lin and Chenfanfu Jiang and Chuang Gan},
      journal={ICLR},
      year={2023},
}

@article{xie2023physgaussian,
      title={PhysGaussian: Physics-Integrated 3D Gaussians for Generative Dynamics}, 
      author={Xie, Tianyi and Zong, Zeshun and Qiu, Yuxing and Li, Xuan and Feng, Yutao and Yang, Yin and Jiang, Chenfanfu},
      journal={CVPR},
      year={2024},
}

@inproceedings{zhang2024physdreamer,
    title={{PhysDreamer}: Physics-Based Interaction with 3D Objects via Video Generation},
    author={Tianyuan Zhang and Hong-Xing Yu and Rundi Wu and Brandon Y. Feng and Changxi Zheng and Noah Snavely and Jiajun Wu and William T. Freeman},
    booktitle={ECCV},
    year={2024},
    organization={Springer}
}

@InProceedings{Abou-Chakra_2024_WACV,
    author    = {Abou-Chakra, Jad and Dayoub, Feras and S\"underhauf, Niko},
    title     = {ParticleNeRF: A Particle-Based Encoding for Online Neural Radiance Fields},
    booktitle = {WACV},
    month     = {January},
    year      = {2024},
    pages     = {5975-5984}
}

@inproceedings{abouchakra-embodiedgaussians,
    title={Physically Embodied Gaussian Splatting: A Realtime Correctable World Model for Robotics},
    author={Jad Abou-Chakra and Krishan Rana and Feras Dayoub and Niko Suenderhauf},
    booktitle={8th Annual Conference on Robot Learning},
    year={2024},
    url={https://openreview.net/forum?id=AEq0onGrN2}
}

@misc{abouchakra2025realissim,
      title={Real-is-Sim: Bridging the Sim-to-Real Gap with a Dynamic Digital Twin}, 
      author={Jad Abou-Chakra and Lingfeng Sun and Krishan Rana and Brandon May and Karl Schmeckpeper and Niko Suenderhauf and Maria Vittoria Minniti and Laura Herlant},
      year={2025},
      eprint={2504.03597},
      archivePrefix={arXiv},
      primaryClass={cs.RO},
      url={https://arxiv.org/abs/2504.03597}, 
}

@InProceedings{Feng_2024_CVPR,
    author    = {Feng, Yutao and Shang, Yintong and Li, Xuan and Shao, Tianjia and Jiang, Chenfanfu and Yang, Yin},
    title     = {PIE-NeRF: Physics-based Interactive Elastodynamics with NeRF},
    booktitle = {Proceedings of the IEEE/CVF Conference on Computer Vision and Pattern Recognition (CVPR)},
    month     = {June},
    year      = {2024},
    pages     = {4450-4461}
}

@article{zhong2024springgaus,
        title     = {Reconstruction and Simulation of Elastic Objects with Spring-Mass 3D Gaussians},
        author    = {Zhong, Licheng and Yu, Hong-Xing and Wu, Jiajun and Li, Yunzhu},
        journal   = {ECCV},
        year      = {2024}
    }

@article{jiang2025phystwin,
            title={PhysTwin: Physics-Informed Reconstruction and Simulation of Deformable Objects from Videos},
            author={Jiang, Hanxiao and Hsu, Hao-Yu and Zhang, Kaifeng and Yu, Hsin-Ni and Wang, Shenlong and Li, Yunzhu},
            journal={ICCV},
            year={2025}
}

@inproceedings{zhang2024dynamics,
    title={Dynamic 3D Gaussian Tracking for Graph-Based Neural Dynamics Modeling},
    author={Zhang, Mingtong and Zhang, Kaifeng and Li, Yunzhu},
    booktitle={8th Annual Conference on Robot Learning},
    year={2024}
}

@inproceedings{zhang2024particle,
  title={Particle-Grid Neural Dynamics for Learning Deformable Object Models from RGB-D Videos},
  author={Zhang, Kaifeng and Li, Baoyu and Hauser, Kris and Li, Yunzhu},
  booktitle={Proceedings of Robotics: Science and Systems (RSS)},
  year={2025}
}

@inproceedings{zhang2024adaptigraph,
      title={AdaptiGraph: Material-Adaptive Graph-Based Neural Dynamics for Robotic Manipulation},
      author={Zhang, Kaifeng and Li, Baoyu and Hauser, Kris and Li, Yunzhu},
      booktitle={Proceedings of Robotics: Science and Systems (RSS)},
      year={2024}
}

@inproceedings{huang2025ParticleFormer,
    title={ParticleFormer: A 3D Point Cloud World Model for Multi-Object, Multi-Material Robotic Manipulation}, 
    author={Suning Huang and Qianzhong Chen and Xiaohan Zhang and Jiankai Sun and Mac Schwager},
    booktitle={9th Conference on Robot Learning (CoRL 2025)},
    year={2025},
}

@misc{obrist2025pokeflex,
      title={PokeFlex: A Real-World Dataset of Volumetric Deformable Objects for Robotics}, 
      author={Jan Obrist and Miguel Zamora and Hehui Zheng and Ronan Hinchet and Firat Ozdemir and Juan Zarate and Robert K. Katzschmann and Stelian Coros},
      year={2025},
      eprint={2410.07688},
      archivePrefix={arXiv},
      primaryClass={cs.RO},
      url={https://arxiv.org/abs/2410.07688}, 
}

@incollection{jiang2016material,
  title={The material point method for simulating continuum materials},
  author={Jiang, Chenfanfu and Schroeder, Craig and Teran, Joseph and Stomakhin, Alexey and Selle, Andrew},
  booktitle={Acm siggraph 2016 courses},
  pages={1--52},
  year={2016}
}

@inproceedings{zong2023neural,
  title={Neural stress fields for reduced-order elastoplasticity and fracture},
  author={Zong, Zeshun and Li, Xuan and Li, Minchen and Chiaramonte, Maurizio M and Matusik, Wojciech and Grinspun, Eitan and Carlberg, Kevin and Jiang, Chenfanfu and Chen, Peter Yichen},
  booktitle={SIGGRAPH Asia 2023 Conference Papers},
  pages={1--11},
  year={2023}
}

@inproceedings{xu2023differentiable,
      title={Differentiable Fluid Physics Parameter Identification By Stirring and For Stirring},
      author={Xu, Wenqiang and Zheng, Dongzhe and Li, Yutong and Ren, Jieji and Lu, Cewu},
      booktitle={2024 IEEE/RSJ International Conference on Intelligent Robots and Systems (IROS)},
      year={2024},
      organization={IEEE}
    }

@inproceedings{Yang2025Differentiable,
  title={Differentiable Physics-based System Identification for Robotic Manipulation of Elastoplastic Materials},
  author={Xintong Yang and Ze Ji and Yu-Kun Lai},
  booktitle={IJRR},
  year={2025}
}

@misc{ravi2024sam2segmentimages,
      title={SAM 2: Segment Anything in Images and Videos}, 
      author={Nikhila Ravi and Valentin Gabeur and Yuan-Ting Hu and Ronghang Hu and Chaitanya Ryali and Tengyu Ma and Haitham Khedr and Roman Rädle and Chloe Rolland and Laura Gustafson and Eric Mintun and Junting Pan and Kalyan Vasudev Alwala and Nicolas Carion and Chao-Yuan Wu and Ross Girshick and Piotr Dollár and Christoph Feichtenhofer},
      year={2024},
      eprint={2408.00714},
      archivePrefix={arXiv},
      primaryClass={cs.CV},
      url={https://arxiv.org/abs/2408.00714}, 
}

@InProceedings{karaev2024cotracker3,
    author    = {Nikita Karaev and Iurii Makarov and Jianyuan Wang and Natalia Neverova and Andrea Vedaldi and Christian Rupprecht},
    title     = {{CoTracker3}: Simpler and Better Point Tracking by Pseudo-Labelling Real Videos},
    journal   = {arxiv},
    year      = {2024}
  }

@misc{warp2022,
  title        = {Warp: A High-performance Python Framework for GPU Simulation and Graphics},
  author       = {Miles Macklin},
  month        = {March},
  year         = {2022},
  note         = {NVIDIA GPU Technology Conference (GTC)},
  howpublished = {\url{https://github.com/nvidia/warp}}
}

@article{li2024_MegaSaM,
  title     = {{MegaSaM}: Accurate, Fast and Robust Structure and Motion from Casual Dynamic Videos},
  author    = {Li, Zhengqi and Tucker, Richard and Cole, Forrester and Wang, Qianqian and Jin, Linyi and Ye, Vickie and Kanazawa, Angjoo and Holynski, Aleksander and Snavely, Noah},
  journal   = {CVPR},
  year      = {2025}
}

@inproceedings{hu2019chainqueen,
  title={Chainqueen: A real-time differentiable physical simulator for soft robotics},
  author={Hu, Yuanming and Liu, Jiancheng and Spielberg, Andrew and Tenenbaum, Joshua B and Freeman, William T and Wu, Jiajun and Rus, Daniela and Matusik, Wojciech},
  booktitle={2019 International conference on robotics and automation (ICRA)},
  pages={6265--6271},
  year={2019},
  organization={IEEE}
}

@article{hu2019difftaichi,
  title={Difftaichi: Differentiable programming for physical simulation},
  author={Hu, Yuanming and Anderson, Luke and Li, Tzu-Mao and Sun, Qi and Carr, Nathan and Ragan-Kelley, Jonathan and Durand, Fr{\'e}do},
  journal={arXiv preprint arXiv:1910.00935},
  year={2019}
}

@article{jiang2015affine,
  title={The affine particle-in-cell method},
  author={Jiang, Chenfanfu and Schroeder, Craig and Selle, Andrew and Teran, Joseph and Stomakhin, Alexey},
  journal={ACM Transactions on Graphics (TOG)},
  volume={34},
  number={4},
  pages={1--10},
  year={2015},
  publisher={ACM New York, NY, USA}
}

@article{su2023generalized,
  title={A generalized constitutive model for versatile MPM simulation and inverse learning with differentiable physics},
  author={Su, Haozhe and Li, Xuan and Xue, Tao and Jiang, Chenfanfu and Aanjaneya, Mridul},
  journal={Proceedings of the ACM on Computer Graphics and Interactive Techniques},
  volume={6},
  number={3},
  pages={1--20},
  year={2023},
  publisher={ACM New York, NY, USA}
}

@article{hu2019taichi,
  title={Taichi: a language for high-performance computation on spatially sparse data structures},
  author={Hu, Yuanming and Li, Tzu-Mao and Anderson, Luke and Ragan-Kelley, Jonathan and Durand, Fr{\'e}do},
  journal={ACM Transactions on Graphics (TOG)},
  volume={38},
  number={6},
  pages={1--16},
  year={2019},
  publisher={ACM New York, NY, USA}
}

@misc{ye2024gsplat,
      title={gsplat: An Open-Source Library for Gaussian Splatting}, 
      author={Vickie Ye and Ruilong Li and Justin Kerr and Matias Turkulainen and Brent Yi and Zhuoyang Pan and Otto Seiskari and Jianbo Ye and Jeffrey Hu and Matthew Tancik and Angjoo Kanazawa},
      year={2024},
      eprint={2409.06765},
      archivePrefix={arXiv},
      primaryClass={cs.CV},
      url={https://arxiv.org/abs/2409.06765}, 
}
}

\end{document}